\documentclass[conference]{IEEEtran}
\IEEEoverridecommandlockouts
\usepackage{cite}
\usepackage{amsmath,amssymb,amsfonts}
\usepackage{graphicx}
\usepackage{textcomp}
\usepackage{amsmath}
\usepackage{algorithm}
\usepackage{algpseudocode}
\usepackage{graphicx}
\usepackage{subcaption}
\usepackage{lipsum}
\usepackage{multicol}
\usepackage{cite}
\usepackage{hyperref}
\usepackage{xcolor}
\usepackage{wrapfig}
\usepackage{blindtext}
\graphicspath{{images/}}
\def\BibTeX{{\rm B\kern-.05em{\sc i\kern-.025em b}\kern-.08em
    T\kern-.1667em\lower.7ex\hbox{E}\kern-.125emX}}

\begin{document}

\title{GA+DDPG+HER: Genetic Algorithm-Based Function Optimizer in Deep Reinforcement Learning for Robotic Manipulation Tasks\\
}

\author{
    \IEEEauthorblockN{
        Adarsh Sehgal\IEEEauthorrefmark{1}\IEEEauthorrefmark{2}, 
        Nicholas Ward\IEEEauthorrefmark{1}\IEEEauthorrefmark{2},  
        Hung Manh La\IEEEauthorrefmark{1}\IEEEauthorrefmark{2},
        Christos Papachristos\IEEEauthorrefmark{2}, and
        Sushil Louis\IEEEauthorrefmark{2}}
    \IEEEauthorblockA{
        \begin{tabular}{cc}
            \begin{tabular}{@{}c@{}}
                \IEEEauthorrefmark{1}
                    Advanced Robotics and Automation (ARA) Laboratory \\
                \IEEEauthorrefmark{2}
                    Department of Computer Science and Engineering, 
                    University of Nevada, Reno, 89557, NV, USA \\
            \end{tabular}
        \end{tabular}
    }
}

\maketitle

\begin{abstract}
Agents can base decisions made using reinforcement learning (RL) on a reward function. The selection of values for the learning algorithm parameters can, nevertheless, have a substantial impact on the overall learning process. In order to discover values for the learning parameters that are close to optimal, we extended our previously proposed genetic algorithm-based Deep Deterministic Policy Gradient and Hindsight Experience Replay approach (referred to as GA+DDPG+HER) in this study. On the robotic manipulation tasks of FetchReach, FetchSlide, FetchPush, FetchPick\&Place, and DoorOpening, we applied the GA+DDPG+HER methodology. Our technique GA+DDPG+HER was also used in the AuboReach environment with a few adjustments. Our experimental analysis demonstrates that our method produces performance that is noticeably better and occurs faster than the original algorithm.
We also offer proof that GA+DDPG+HER beat the current approaches. The final results support our assertion and offer sufficient proof that automating the parameter tuning procedure is crucial and does cut down learning time by as much as $57\%$.
\end{abstract}

\begin{IEEEkeywords}
DRL, DDPG+HER, Reinforcement Learning, Genetic Algorithm, GA+DDPG+HER, DDPG, HER
\end{IEEEkeywords}
\vspace{-12pt}
\section{Introduction}
Recently, Reinforcement Learning (RL) has been put to significant uses such as standard robotic manipulation \cite{sehgal2022aacher}. Each of these applications make use of RL as an encouraging substitute to automating manual effort.  

In this paper, we are specifically using DDPG \cite{lillicrap2015continuous} combined with HER \cite{andrychowicz2017hindsight} to train Deep Reinforcement Learning (DRL) policies. Our primary contributions are as follows: 1) For algorithm analysis, we deployed our own previously developed GA+DDPG+HER algorithm \cite{sehgal2019deep} on Aubo-i5 simulated and real environments. 2) As the GA develops, the training process is examined utilizing a variety of criteria. 3) GA+DDPG+HER was compared to existing techniques. Open source code is available at \textcolor{orange}{\href{https://github.com/aralab-unr/ga-drl-aubo-ara-lab}{https://github.com/aralab-unr/ga-drl-aubo-ara-lab}}.

The algorithm GA+DDPG+HER was discovered in our earlier work \cite{sehgal2019deep} and \cite{sehgal2022automatic}. \cite{sehgal2022gaddpgher} provides a thorough background, experimental conditions, implementation details, training evaluation, and analysis of this research.
Some of the closely related work includes \cite{10.1007/978-3-030-33723-0_29}. The results from these papers provide more evidence that when a GA is used to automatically tune the hyper-parameters for DDPG+HER, efficiency can be largely improved. The difference can greatly impact the time it takes for a learning agent to learn.

\section{Experimental Results}\label{chapter_four}

\begin{figure}
\centering
\begin{multicols}{2}    \includegraphics[width=0.25\textwidth]{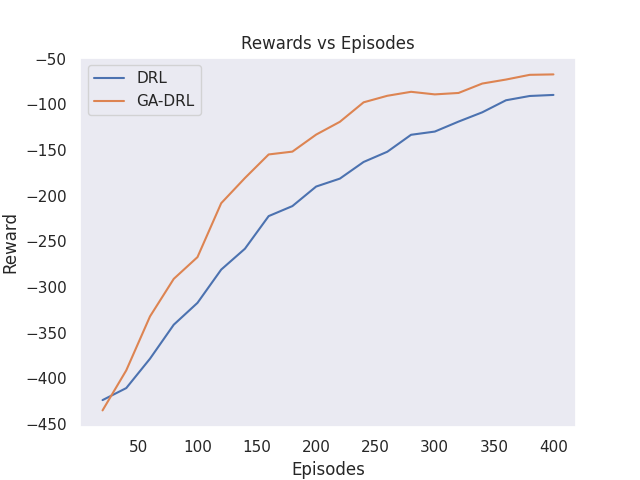}
    \subcaption{AuboReach}
\includegraphics[width=0.25\textwidth]{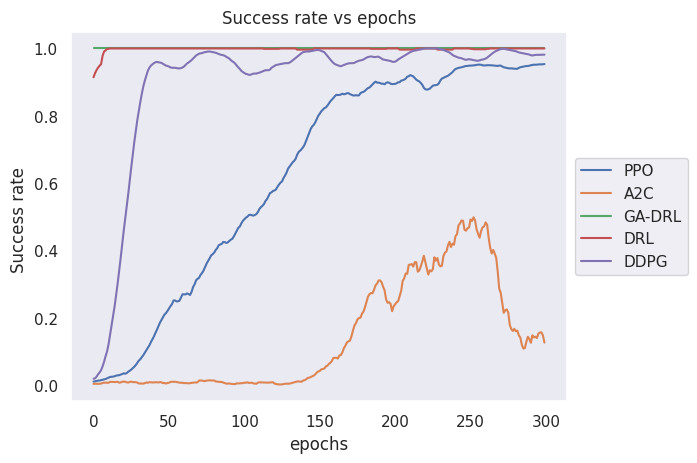}
\subcaption{FetchReach}
\end{multicols}
\vspace{-10pt}
\caption{GA+DDPG+HER vs.. existing algorithm's efficiency evaluation plots (total reward vs.. episodes) when all the six parameters are found by GA. FetchReach is averaged over two runs, and AuboReach is averaged over ten runs. }
\vspace{-10pt}
\label{fig:analysisPlots}
\end{figure}




\subsection{Training evaluation}

\begin{figure*}
\centering
\begin{multicols}{6}
    \includegraphics[width=4.5cm,height=3.8cm]{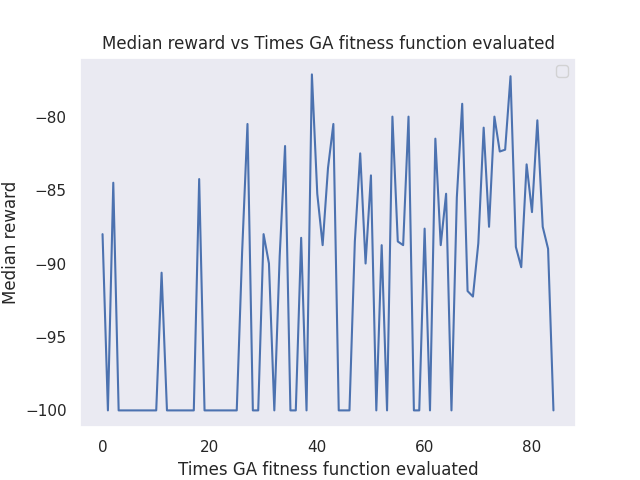}
    \subcaption{FetchPick\&Place - Median reward vs. Times GA fitness function evaluated}
    \includegraphics[width=4.5cm,height=3.8cm]{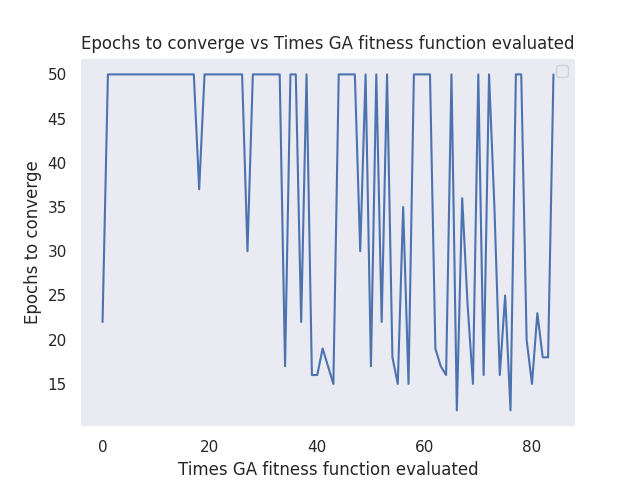}
    \subcaption{FetchPick\&Place - Epochs vs. Times GA fitness function evaluated}
    \includegraphics[width=4.5cm,height=3.8cm]{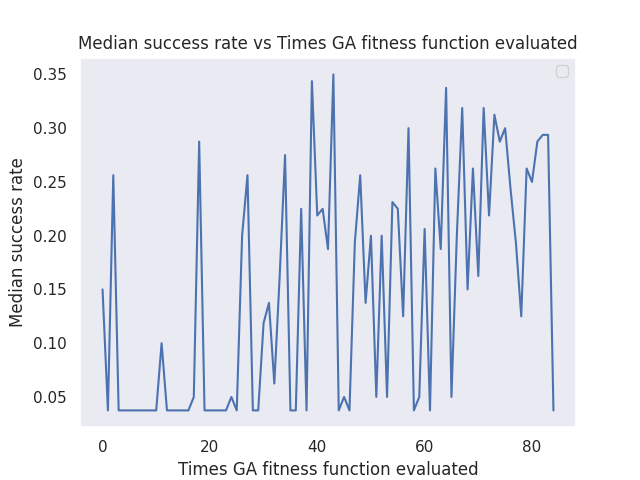}
    \subcaption{FetchPick\&Place - Median success rate vs. Times GA fitness function evaluated}
    \includegraphics[width=4.5cm,height=3.8cm]{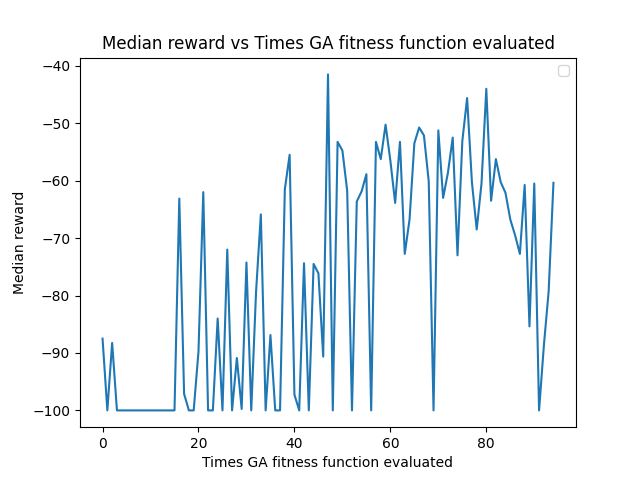}
    \subcaption{FetchPush - Median reward vs. Times GA fitness function evaluated}
    \includegraphics[width=4.5cm,height=3.8cm]{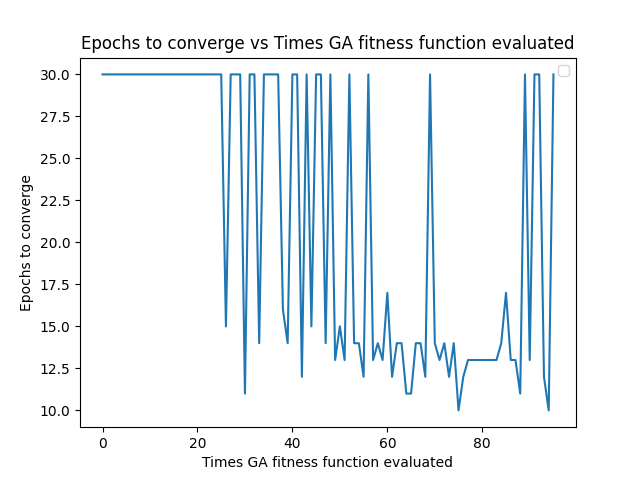}
    \subcaption{FetchPush - Epochs vs. Times GA fitness function evaluated}
    \includegraphics[width=4.5cm,height=3.8cm]{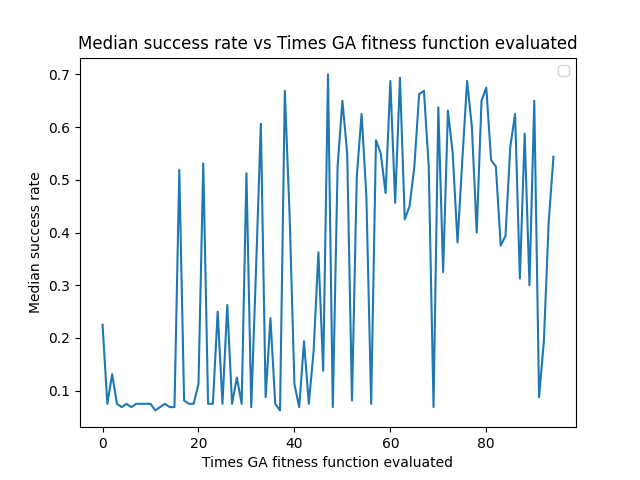}
    \subcaption{FetchPush - Median success rate vs. Times GA fitness function evaluated}
    \includegraphics[width=4.5cm,height=3.8cm]{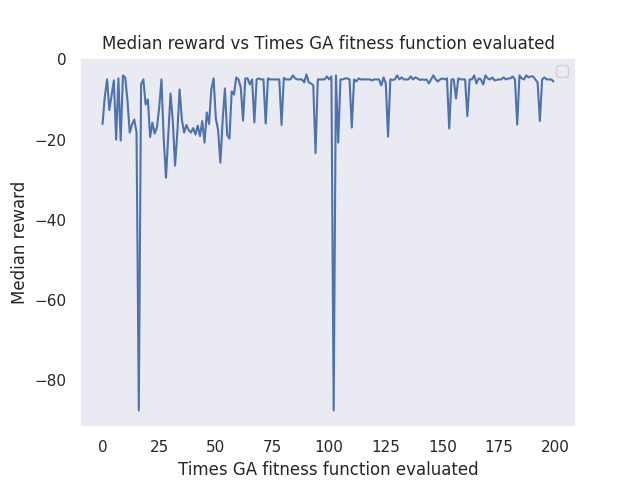}
    \subcaption{FetchReach - Median reward vs. Times GA fitness function evaluated}
    \includegraphics[width=4.5cm,height=3.8cm]{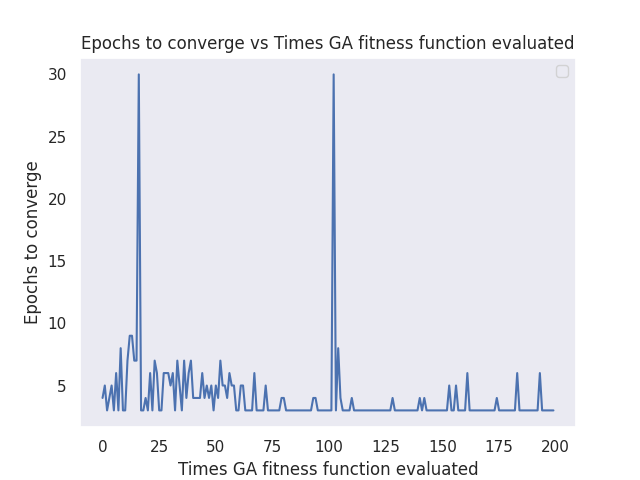}
    \subcaption{FetchReach - Epochs vs. Times GA fitness function evaluated}

    \includegraphics[width=4.5cm,height=3.8cm]{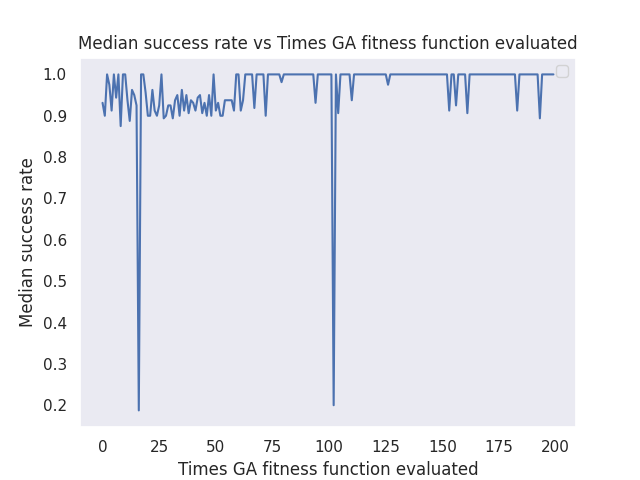}
    \subcaption{FetchReach - Median success rate vs. Times GA fitness function evaluated}
    \includegraphics[width=4.5cm,height=3.8cm]{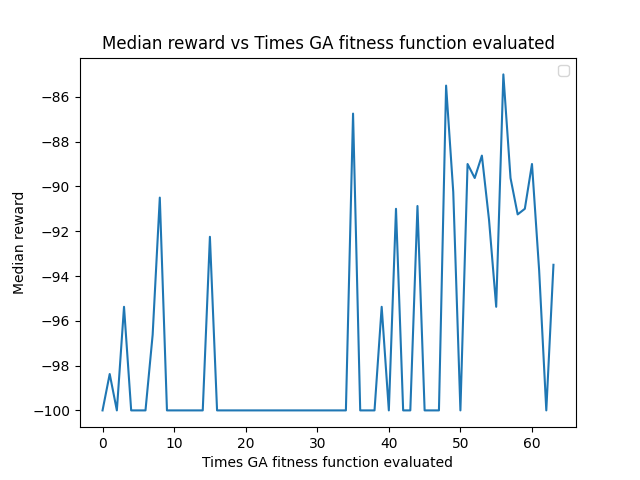}
    \subcaption{FetchSlide - Median reward vs. Times GA fitness function evaluated}
    \includegraphics[width=4.5cm,height=3.8cm]{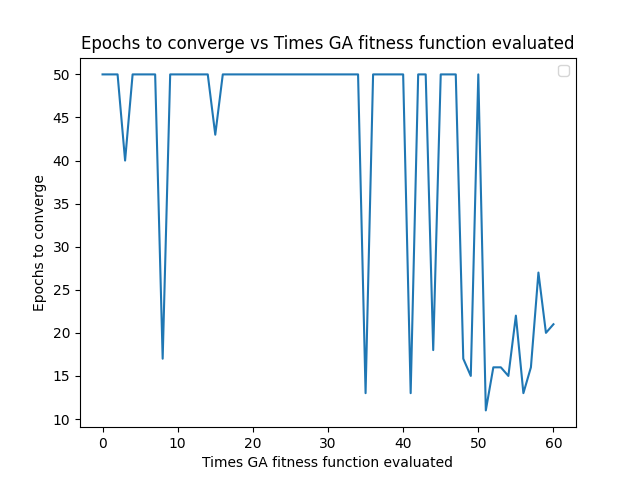}
    \subcaption{FetchSlide - Epochs vs. Times GA fitness function evaluated}
    \includegraphics[width=4.5cm,height=3.8cm]{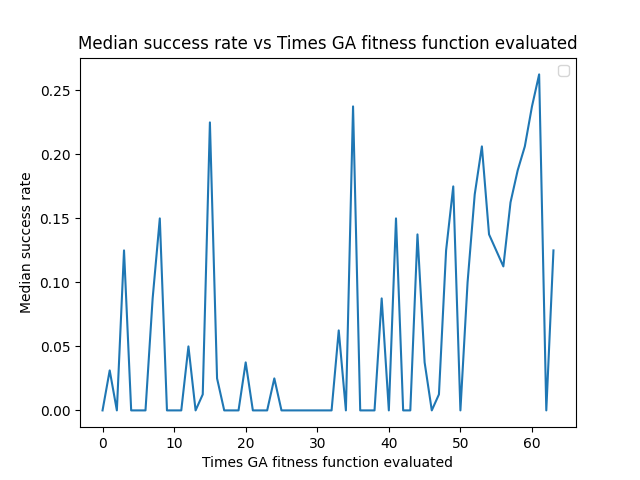}
    \subcaption{FetchSlide - Median success rate vs. Times GA fitness function evaluated}
    \includegraphics[width=4.5cm,height=3.8cm]{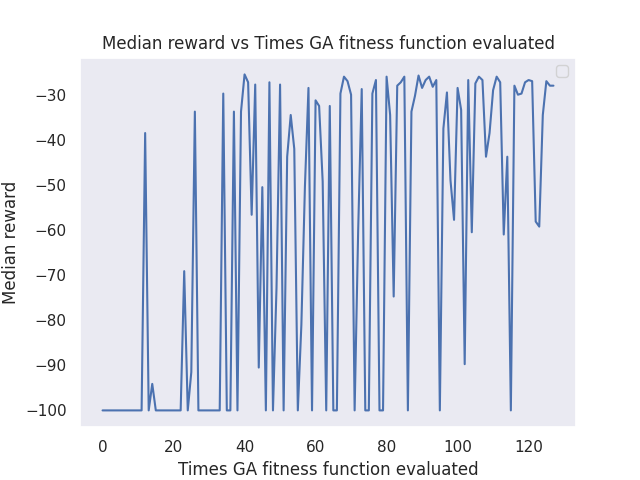}
    \subcaption{DoorOpening - Median reward vs. Times GA fitness function evaluated}
    \includegraphics[width=4.5cm,height=3.8cm]{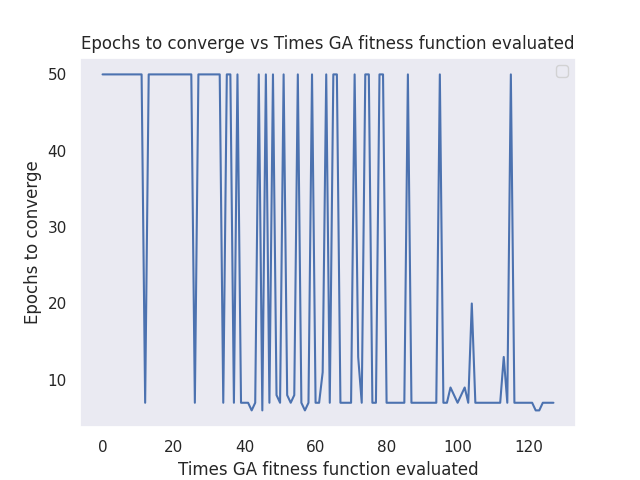}
    \subcaption{DoorOpening - Epochs vs. Times GA fitness function evaluated}
    \includegraphics[width=4.5cm,height=3.8cm]{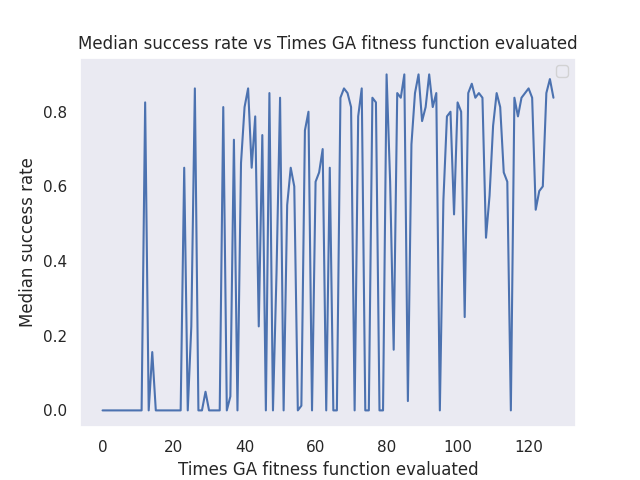}
    \subcaption{DoorOpening - Median success rate vs. Times GA fitness function evaluated}
    \includegraphics[width=3cm,height=3.8cm]{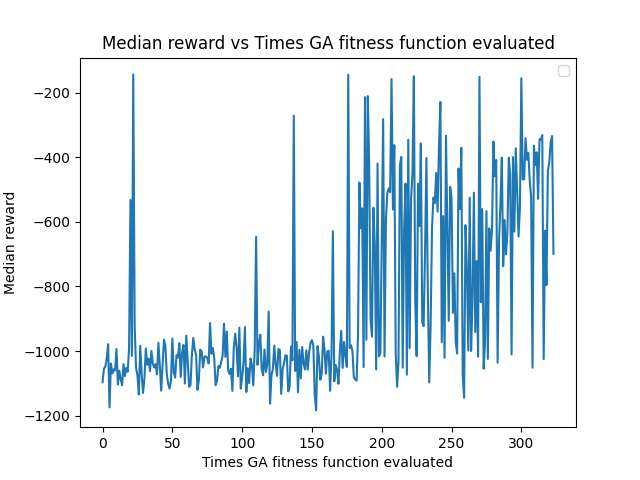}
    \subcaption{AuboReach - Median reward vs. Times GA fitness function evaluated}

    \includegraphics[width=3cm,height=3.8cm]{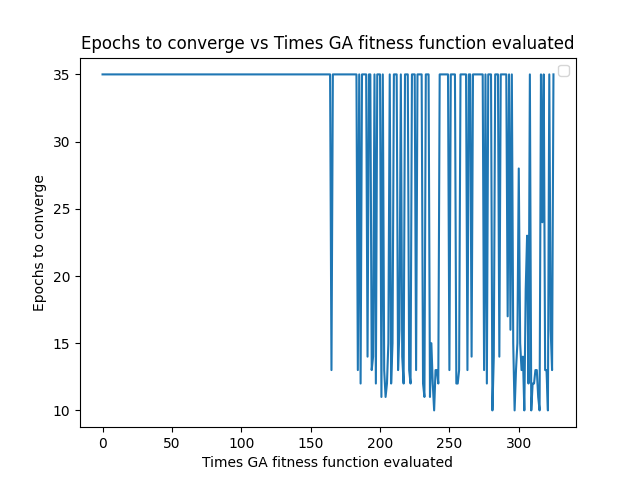}
    \subcaption{AuboReach - Epochs vs. Times GA fitness function evaluated}
    \includegraphics[width=3cm,height=3.8cm]{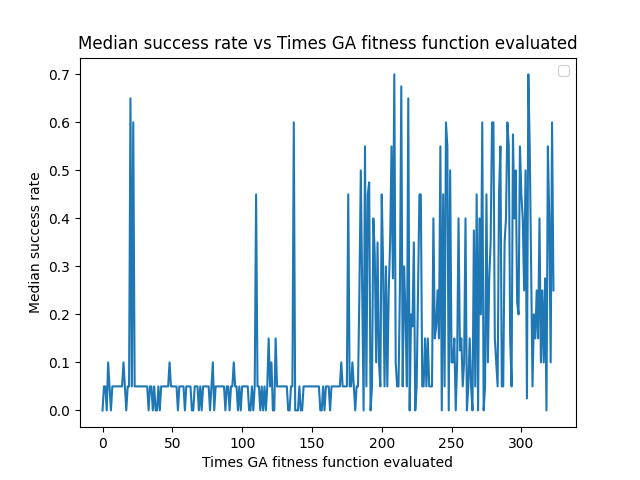}
    \subcaption{AuboReach - Median success rate vs. Times GA fitness function evaluated}
\end{multicols}
\caption{GA+DDPG+HER training evaluation plots, when all the 6 parameters are found by GA (this is for one GA run.)}
\label{fig:gaTrainingEvaluationPlots}
\end{figure*}

Figure \ref{fig:analysisPlots} contrasts DDPG+HER with GA+DDPG+HER, whereas Figure \ref{fig:gaTrainingEvaluationPlots} shows how the system performs better as the GA evolves. 

\vspace{-8pt}
\subsection{Efficiency evaluation}
\vspace{-5pt}
This section compares DDPG+HER to GA+DDPG+HER in a variety of environmental settings. The most effective algorithm is indicated in each of the tables \ref{table:averageepisodes}-\ref{table:averageepochs} by a number in bold.

\begin{table}[ht]
\tiny
\centering
\begin{tabular}{|p{0.13\linewidth}|p{0.117\linewidth}|p{0.07\linewidth}|p{0.072\linewidth}|p{0.07\linewidth}|p{0.09\linewidth}|p{0.075\linewidth}|} 
 \hline
 Method & FetchPick\&Place & FetchPush & FetchReach & FetchSlide & DoorOpening & AuboReach\\ 
 \hline
 DDPG+HER & 6,000 & 2,760 & 100 & \textbf{4,380} & 960 & 320\\ 
 GA+DDPG+HER & \textbf{2,270} & \textbf{1,260} & \textbf{60} & 6,000 & \textbf{180} & \textbf{228}\\
 \hline
\end{tabular}
\caption{Efficiency evaluation: Average (over ten runs) episodes comparison to reach the goal, for all the tasks.}
\label{table:averageepisodes}
\end{table}




\vspace{-15pt}
\begin{table}[ht]
\tiny
\centering
\begin{tabular}{|p{0.13\linewidth}|p{0.117\linewidth}|p{0.07\linewidth}|p{0.072\linewidth}|p{0.07\linewidth}|p{0.09\linewidth}|p{0.075\linewidth}|} 
 \hline
 Method & FetchPick\&Place & FetchPush & FetchReach & FetchSlide & DoorOpening & AuboReach\\ [0.5ex] 
 \hline
 DDPG+HER & 3069.981 & 1314.477 & 47.223 & \textbf{2012.645} & 897.816 & 93.258\\ 
 GA+DDPG+HER & \textbf{1224.697} & \textbf{565.178} & \textbf{28.028} & 3063.599 & \textbf{167.883} & \textbf{66.818}\\[1ex] 
 \hline
\end{tabular}
\caption{Efficiency evaluation: Average (over 10 runs) running time (s) comparison to reach the goal, for all the tasks.}
\label{table:averagerunningtime}
\end{table}

\vspace{-15pt}
\begin{table}[ht]
\tiny
\centering
\begin{tabular}{|p{0.13\linewidth}|p{0.117\linewidth}|p{0.07\linewidth}|p{0.072\linewidth}|p{0.07\linewidth}|p{0.09\linewidth}|p{0.075\linewidth}|} 
 \hline
 Method & FetchPick\&Place & FetchPush & FetchReach & FetchSlide & DoorOpening & AuboReach\\ [0.5ex] 
 \hline
 DDPG+HER & 300,000 & 138,000 & 5000 & \textbf{219,000} & 48000 & 65,600\\ 
 GA+DDPG+HER & \textbf{113,000} & \textbf{63,000} & \textbf{3000} & 300,000 & \textbf{9000} & \textbf{46,000}\\[1ex] 
 \hline
\end{tabular}
\caption{Efficiency evaluation: Average (over ten runs) steps comparison to reach the goal, for all the tasks.}
\label{table:averagesteps}
\end{table}

\vspace{-12pt}
\begin{table}[ht]
\tiny
\centering
\begin{tabular}{|p{0.13\linewidth}|p{0.117\linewidth}|p{0.07\linewidth}|p{0.072\linewidth}|p{0.07\linewidth}|p{0.09\linewidth}|p{0.075\linewidth}|} 
 \hline
 Method & FetchPick\&Place & FetchPush & FetchReach & FetchSlide & DoorOpening & AuboReach\\ [0.5ex] 
 \hline
 DDPG+HER & 60 & 27.6 & 5 & \textbf{43.8} & 47 & 16\\ 
 GA+DDPG+HER & \textbf{22.6} & \textbf{12.6} & \textbf{3} & 60 & \textbf{8} & \textbf{11.4}\\[1ex] 
 \hline
\end{tabular}
\caption{Efficiency evaluation: Average (over 10 runs) epochs comparison to reach the goal, for all the tasks.}
\vspace{-20pt}
\label{table:averageepochs}
\end{table}

\bibliography{root.bib}
\bibliographystyle{IEEEtran}

\end{document}